%% file: arxiv-metasel.tex
\documentclass[sigconf]{acmart}

\AtBeginDocument{%
  }

\setcopyright{none}
\acmDOI{}
\acmISBN{}

\usepackage{amsmath}
\usepackage{amssymb}
\usepackage{amsfonts}
\usepackage{booktabs}
\usepackage[table]{xcolor}
\usepackage{multirow}
\usepackage{adjustbox}
\usepackage{rotating}
\usepackage{bbm}
\usepackage{algorithm}
\usepackage{algorithmic}
\usepackage{subcaption}
\usepackage{pifont}  % for \ding{51} (checkmark) and \ding{55} (cross)
\usepackage{tikz}
\usetikzlibrary{arrows.meta,positioning,calc,fit,shapes,decorations.pathreplacing,shadows}

\newtheorem{theorem}{Theorem}[section]
\newtheorem{proposition}[theorem]{Proposition}

\title[Meta-Sel for ICL]{Meta-Sel: Efficient Demonstration Selection for In-Context Learning via Supervised Meta-Learning}

\author{Xubin Wang}
\affiliation{%
  \institution{BNU-BNBU Institute of Artificial Intelligence and Future Networks, Beijing Normal-Hong Kong Baptist University and Beijing Normal University at Zhuhai}
  \city{Zhuhai}
  \country{China}
}

\author{Weijia Jia}
\affiliation{%
  \institution{BNU-BNBU Institute of Artificial Intelligence and Future Networks, Beijing Normal-Hong Kong Baptist University and Beijing Normal University at Zhuhai}
  \city{Zhuhai}
  \country{China}
}

\begin{document}

\begin{abstract}
Demonstration selection is a practical bottleneck in in-context learning (ICL): under a tight prompt budget, accuracy can change substantially depending on which few-shot examples are included, yet selection must remain cheap enough to run per query over large candidate pools. We propose \textsc{Meta-Sel}, a lightweight supervised meta-learning approach for intent classification that learns a fast, interpretable scoring function for (candidate, query) pairs from labeled training data.

\textsc{Meta-Sel} constructs a meta-dataset by sampling pairs from the training split and using class agreement as supervision, then trains a calibrated logistic regressor on two inexpensive meta-features: TF--IDF cosine similarity and a length-compatibility ratio. At inference time, the selector performs a single vectorized scoring pass over the full candidate pool and returns the top-$k$ demonstrations, requiring no model fine-tuning, no online exploration, and no additional LLM calls. This yields deterministic rankings and makes the selection mechanism straightforward to audit via interpretable feature weights.

Beyond proposing \textsc{Meta-Sel}, we provide a broad empirical study of demonstration selection, benchmarking \textbf{12} methods---spanning prompt engineering baselines, heuristic selection, reinforcement learning, and influence-based approaches---across four intent datasets and five open-source LLMs. Across this benchmark, \textsc{Meta-Sel} consistently ranks among the top-performing methods, is particularly effective for smaller models where selection quality can partially compensate for limited model capacity, and maintains competitive selection-time overhead.
\end{abstract}




\maketitle

\section{Introduction}

Large language models (LLMs) have become a strong foundation for NLP, and in-context learning (ICL) enables task adaptation by conditioning on a small set of demonstrations in the prompt~\cite{brown2020language,wei2021finetuned,dong2024survey,zhou2024mystery}. As context windows expand, ICL can also be studied in many-shot and long-context regimes, but using more demonstrations further amplifies prompt cost and system-level efficiency constraints~\cite{agarwal2024many,bertsch2025context}. Practical toolkits reflect this growing design space by exposing interchangeable retrieval, prompting, and inference components for ICL pipelines~\cite{wu2023openicl}, while dedicated surveys have catalogued retrieval-based demonstration strategies~\cite{dong2024survey}. Despite rapid progress, ICL remains only partially understood: theoretical accounts characterize ICL as implicit Bayesian inference~\cite{xie2022explanation,zhang2025and}, implicit gradient descent~\cite{dai2023can}, or task identification over latent structure~\cite{wies2023learnability,akyureklearning}, while information-theoretic analyses bound its sample complexity~\cite{jeon2024information}.

In few-shot ICL, the model is conditioned on a small set of demonstrations that specify the task in the prompt, and a well-known practical bottleneck is that performance depends heavily on \emph{which} demonstrations are chosen (and how they are ordered)~\cite{wu2023self,chen2023relation}. Given a query $x_q$ and a candidate pool $\mathcal{D}_{\text{train}}$, the fundamental problem is to select $k$ demonstrations that maximize task accuracy under strict compute and context-length constraints. This selection problem is challenging for at least three reasons. \textbf{First}, the prompt budget is limited: as models scale, longer contexts remain expensive, so $k$ is typically small and the marginal value of each demonstration is high. \textbf{Second}, usefulness is query-dependent: a demonstration that helps one query may be irrelevant or even harmful for another (e.g., due to spurious lexical overlap or label confusion). \textbf{Third}, selection must be efficient: many deployment settings require scoring thousands of candidates per query, ruling out expensive online optimization or repeated LLM calls.

\input{motivation_metasel.tex}

Existing approaches span a broad design space. \textbf{Prompt engineering methods}~\cite{wei2022chain,zhou2023leasttomost,madaan2023self,wu2025enhancing} modify the prompt format or enhance annotations without explicitly selecting examples. \textbf{Heuristic selection methods}~\cite{yang2023representative,margatina2023active} rely on hand-crafted criteria such as similarity, diversity, or uncertainty. \textbf{Reinforcement learning (RL) approaches}~\cite{zhang2022active,wang2024rdes} learn adaptive policies through reward-driven optimization but often incur substantial online exploration cost and variance. \textbf{Information-theoretic techniques}~\cite{settles2009active,liu2023towards} estimate uncertainty or expected information gain to quantify demonstration informativeness, while \textbf{supervised retrieval and meta-learning approaches}~\cite{rubin2022learning,min2022metaicl,sinha2024maml} learn selection signals from data but can introduce additional training complexity. \textbf{Data valuation methods}~\cite{zhou2025valuing} use causal inference to quantify individual training sample contributions, while \textbf{influence-based approaches}~\cite{askari2025unraveling} leverage influence functions for demonstration reweighting. Finally, \textbf{set-level search and subset selection} methods address interactions among demonstrations but may require expensive evaluation or multi-stage filtering~\cite{li2023finding,ye2023compositional,qin2024context}. Despite strong progress, it remains unclear how these families compare across datasets and model scales, and when their additional complexity is justified.

Three gaps persist in practice. \textbf{(1) Lack of systematic comparison.} Most studies evaluate a small subset of methods under narrow settings, making cross-paper conclusions brittle---a gap echoed by recent surveys~\cite{dong2024survey}. \textbf{(2) Unclear selection criteria.} Practitioners face many method choices but limited guidance on what properties (labels, similarity, diversity, uncertainty, causal importance~\cite{zhou2025valuing}) matter most under different budgets and model scales~\cite{shi2024larger}. \textbf{(3) Efficiency--effectiveness trade-offs are under-characterized.} Methods that are accurate in principle may be impractical at test time, while simple baselines may be surprisingly strong~\cite{margatina2023active}.

We address these gaps through a combined empirical and methodological contribution. \textbf{First}, we conduct the most comprehensive evaluation of demonstration selection to date, covering \textbf{12} prompt engineering and example selection methods across \textbf{4} intent classification datasets and \textbf{5} open-source language models (4B--20B parameters), totaling 240+ experimental configurations---substantially broader in both method count and model diversity than prior benchmarks. \textbf{Second}, we propose \textbf{Meta-Sel}, a supervised meta-learning approach that casts example selection as a binary classification problem over (candidate, query) pairs (Figure~\ref{fig:motivation}). Meta-Sel trains offline on lightweight, interpretable meta-features (TF--IDF similarity and a length-compatibility ratio) to predict whether a candidate is likely to share the query label, and then ranks candidates by the predicted probability. \textbf{Third}, our systematic analysis reveals actionable insights---including when advanced selection helps most (smaller models), when it can be harmful (large models on fine-grained label spaces), and how method families compare under controlled conditions---providing practitioners with a decision framework for deploying ICL in production.

Critically, Meta-Sel targets a practical ``sweet spot'' on the efficiency--effectiveness frontier: unlike meta-training the \emph{LLM} itself to improve ICL~\cite{min2022metaicl}, Meta-Sel only learns a tiny selector; unlike learning a prompt retriever that relies on model-based supervision or expensive pair scoring~\cite{rubin2022learning}, Meta-Sel uses features computable from text alone; and unlike multi-stage filter-then-search pipelines that leverage LLM feedback~\cite{li2023finding}, Meta-Sel performs a single deterministic scoring pass at inference time. This design yields fast selection, deterministic rankings, and straightforward auditing via feature weights.

\textbf{Why label agreement is a usable proxy.} In intent classification, demonstrations primarily serve to convey a mapping from input utterances to a discrete label set. Under this setting, selecting demonstrations from the same class as the query is a strong and directly measurable signal of relevance: such examples provide label-consistent phrasing and decision cues that the model can copy or analogize from within the prompt. Meta-Sel operationalizes this idea by constructing meta-labels using ground-truth training labels (whether a candidate and query share the same label) and learning, from meta-features, which candidates are likely to match a query. This makes the supervision signal explicit and verifiable, while keeping inference-time selection lightweight.

\textbf{Key findings.} (1) Meta-Sel consistently ranks among the top methods across model--dataset combinations, matching or exceeding more complex RL and information-theoretic baselines in many settings. (2) Gains are especially pronounced for smaller models, showing that better selection can partially compensate for limited model capacity. (3) Similarity-based retrieval remains a strong baseline, and Meta-Sel can further improve selection by learning how to weight simple, observable signals (e.g., similarity and length) from labeled training data. (4) No single method dominates universally; optimal choices depend on dataset diversity, model scale, and computational budget.

\section{Related Work}

\subsection{In-Context Learning}
ICL has emerged as a transformative paradigm in NLP, enabling models to adapt to new tasks by conditioning on a small set of demonstrations within the input context~\cite{brown2020language,wei2021finetuned,liu2023pre}. Comprehensive surveys summarize rapid progress and open questions across theory, prompting, and evaluation~\cite{dong2024survey,zhou2024mystery}. On the theoretical side, ICL has been analyzed through multiple complementary lenses: implicit Bayesian inference~\cite{xie2022explanation,zhang2025and}, learnability frameworks~\cite{wies2023learnability}, information-theoretic bounds on sample complexity~\cite{jeon2024information}, and implicit gradient descent within forward passes~\cite{dai2023can}. Recent work further highlights that larger models exploit ICL differently from smaller ones, with scale inducing qualitatively distinct learning strategies~\cite{shi2024larger}. On the empirical side, the role of demonstrations is subtle: ground-truth labels in demonstrations are not always strictly necessary, yet the \emph{choice} and \emph{format} of demonstrations can strongly affect performance~\cite{min2022rethinking}, and ICL predictions exhibit a strong negative correlation between sensitivity to perturbations and accuracy~\cite{chen2023relation}.

\subsection{Demonstration Selection Techniques}
Effective demonstration selection is crucial for ICL success. Traditional approaches rely on heuristics or statistical measures: uncertainty-based selection draws on active learning principles~\cite{settles2009active,margatina2023active}, while diversity-based methods ensure that the selected set covers a broad range of the input space~\cite{levy2023diverse}. Coverage-based techniques, such as BERTScore-Recall, further promote balanced representation~\cite{gupta2023coverage}, and skill-based methods like Skill-KNN optimize selection by filtering irrelevant features~\cite{an2023skill}. Representative sampling~\cite{yang2023representative} focuses on static diversity, while uncertainty-driven selection~\cite{mavromatis2023examples} ranks candidates without training a policy.

Beyond heuristics, a rich line of recent work learns selectors or retrievers from supervision. Supervised prompt retrieval trains dense retrievers for selecting demonstrations~\cite{rubin2022learning}. A Bayesian perspective frames LLMs as latent variable models, enabling principled example selection that transfers across model scales~\cite{wang2023large}. Information-gain--based approaches maximize the informativeness of selected demonstrations~\cite{liu2023towards}, while data-valuation methods quantify individual training sample contributions through causal inference~\cite{zhou2025valuing}. Iterative demonstration selection alternates between reasoning and retrieval to find context-dependent examples~\cite{qin2024context}. Influence-function--based methods reweight demonstrations by estimating their impact on predictions, and have recently been extended to noisy and multi-task ICL settings~\cite{askari2025unraveling}. Set-level optimization incorporates interactions among demonstrations via LLM feedback~\cite{li2023finding}.

\subsection{Meta-Learning for Example Selection}
Meta-learning, or ``learning to learn,'' offers a principled approach to example selection by learning generalizable patterns across tasks~\cite{finn2017model,nichol2018first}. Early work on in-context tuning recasts task adaptation as sequence prediction and meta-trains language models to learn from in-context examples~\cite{chen2022meta}. MetaICL~\cite{min2022metaicl} extends this idea by meta-training language models on diverse task collections to become better in-context learners, while MAML-en-LLM~\cite{sinha2024maml} applies model-agnostic meta-learning directly to LLMs, achieving improved in-context generalization on both seen and unseen domains---demonstrating the effectiveness of gradient-based meta-training at the KDD scale. In contrast to these approaches that meta-train the \emph{model itself}, our work formulates \emph{selection} as a supervised meta-learning problem: a lightweight classifier learns to predict example usefulness from cheap meta-features. This differs from gradient-based meta-learning like MAML~\cite{finn2017model} by avoiding iterative inner-loop optimization, and from RL-based selection~\cite{scarlatos2023reticl,zhang2022active} by eliminating online exploration. Compared with neural meta-selectors, our TF--IDF + logistic regression scorer is intentionally simple to keep inference-time selection fast and auditable.

\subsection{Supervised Scoring and Retrieval for ICL}
Meta-Sel is also related to retrieval-style demonstration selection, where candidates are ranked by a score computed from (query, candidate) features. Similarity retrieval provides an efficient and often strong baseline, but it uses a fixed heuristic and does not exploit supervision from labeled training data. In contrast, Meta-Sel learns an \emph{offline} scoring function that maps simple, observable pair features to an estimate of label agreement, which is then used for ranking. This connects Meta-Sel to a broader family of supervised scoring approaches---including data-valuation methods that quantify per-sample contributions via causal inference~\cite{zhou2025valuing} and influence-function--based reweighting~\cite{askari2025unraveling}---that learn to map meta-features into usefulness estimates while remaining strictly inference-time lightweight (single-pass scoring over the pool) and avoiding repeated LLM calls.

A key distinction is that many learned retrievers optimize dense representations and rely on model-based supervision for pair labeling or scoring~\cite{rubin2022learning}, whereas LLM-aware selectors require the target model in the loop~\cite{wang2023large}. Meta-Sel instead learns a calibrated, interpretable scorer over inexpensive TF--IDF statistics, allowing deterministic rankings with minimal infrastructure. Robustness to demonstration sensitivity---a concern highlighted by recent empirical studies~\cite{chen2023relation,he2025data}---is further supported by Meta-Sel's deterministic and auditable selection mechanism.

\section{Methods}

% Framework overview figure
\input{metasel_framework_new.tex}

\subsection{Meta-Sel: Supervised Meta-Learning for Example Selection}

\subsubsection{Problem Formulation and Theoretical Framework}

Meta-Sel formulates example selection as a supervised classification problem at the meta-level. Given a labeled training set $\mathcal{D}_{\text{train}} = \{(x_i, y_i)\}_{i=1}^{N}$, we build a supervised \emph{meta}-dataset whose examples are (query, candidate) pairs from the training split. In our implementation, we construct $\mathcal{M}$ by sampling a set of meta-queries and meta-candidates from $\mathcal{D}_{\text{train}}$ (to control cost) and labeling each pair by whether the two examples share the same ground-truth class:

\begin{equation}
\mathcal{M} = \{(f(x_c, x_q), \ell_{cq}) \mid x_q \in \mathcal{Q},\; x_c \in \mathcal{C}\},
\quad \ell_{cq}=\mathbbm{1}[y_c = y_q]
\end{equation}

where $\mathcal{Q}\subseteq\mathcal{D}_{\text{train}}$ is a random sample of meta-queries and $\mathcal{C}\subseteq\mathcal{D}_{\text{train}}$ is a random sample of meta-candidates (defaults: $|\mathcal{Q}|=60$, $|\mathcal{C}|=300$). This construction uses training labels only and avoids test leakage.

\textbf{Feature Engineering.} Our implementation uses two lightweight meta-features:

\begin{itemize}
\item \textbf{Cosine similarity}: cosine similarity between TF-IDF vectors of the query and candidate texts:
\begin{equation}
\text{sim}(x_c, x_q) = \frac{\phi(x_c)^\top \phi(x_q)}{\|\phi(x_c)\|_2 \|\phi(x_q)\|_2}
\end{equation}
where $\phi(\cdot)$ is a TF-IDF vectorizer (we use \texttt{TfidfVectorizer(stop\_words='english')} from scikit-learn). For efficiency at test time, the TF-IDF vectorizer is fit once on all training texts and similarities are computed in a vectorized manner.

\item \textbf{Length ratio}: a simple length compatibility feature,
\begin{equation}
r(x_c, x_q) = \frac{|x_c|}{\max(1, |x_q|)}
\end{equation}
where $|\cdot|$ is the text length as implemented (string length). This discourages selecting candidates that are disproportionately long relative to the query.
\end{itemize}

The final feature vector is $f(x_c, x_q) = [\text{sim}(x_c, x_q), r(x_c, x_q)]^\top \in \mathbb{R}^2$. While more sophisticated feature selection techniques exist for high-dimensional settings~\cite{wang2024mel,wang2022self}, our deliberately minimal feature set keeps inference fast and interpretable.

\subsubsection{Meta-Learner Training and Selection}

We train a logistic regression classifier $h_\theta: \mathbb{R}^2 \rightarrow [0,1]$ on $\mathcal{M}$ to predict whether a (candidate, query) pair shares the same label:

\begin{equation}
h_\theta(f) = \sigma(\theta^\top f + b) = \frac{1}{1 + \exp(-\theta^\top f - b)}
\end{equation}

where $\theta \in \mathbb{R}^2$ are feature weights and $b$ is the bias. In code, we use \texttt{sklearn} \texttt{LogisticRegression} with \texttt{class\_weight='balanced'} and \texttt{max\_iter=200} to mitigate class imbalance.

\textbf{Test-time selection.} For a test query $x_{\text{test}}$, we score each training candidate $x_c \in \mathcal{D}_{\text{train}}$ using the learned classifier on the two features:

\begin{equation}
s_c = h_\theta\big([\text{sim}(x_c, x_{\text{test}}),\; r(x_c, x_{\text{test}})]\big).
\end{equation}

We then select the top-$k$ candidates with the highest scores:

\begin{equation}
\mathcal{E}(x_{\text{test}}) = \text{top-}k(\{(x_c, s_c) \mid x_c \in \mathcal{D}_{\text{train}}\})
\end{equation}

\subsubsection{Algorithm Summary}
The full pseudocode is given in Algorithm~\ref{alg:metasel} (Appendix~\ref{app:algorithm}). The key property for deployment is that all learning happens offline on training data, while test-time selection is a single vectorized scoring pass over the candidate pool.

\subsubsection{Theoretical Justification}

\textbf{Utility-based ranking objective.} For intent classification, a natural notion of demonstration ``usefulness'' is whether the candidate shares the query label. Define the per-pair utility
\begin{equation}
u(x_c, x_q) = \mathbbm{1}[y_c = y_q].
\end{equation}
For a fixed query $x_q$ and a selector that returns a size-$k$ set $S\subseteq\mathcal{D}_{\text{train}}$, the expected number of label-matching demonstrations is
\begin{equation}
\mathbb{E}\big[\sum_{x_c\in S} u(x_c, x_q)\,\big|\,x_q\big] = \sum_{x_c\in S} P(y_c=y_q\mid x_c, x_q).
\end{equation}
Thus, the Bayes-optimal ranking scores candidates by the conditional probability $P(y_c=y_q\mid x_c, x_q)$. Meta-Sel approximates this quantity with a probabilistic classifier $h_\theta(f(x_c,x_q))$ and then selects the top-$k$ candidates by this score.

\textbf{Surrogate learning with a proper scoring rule.} The implemented meta-learner is logistic regression trained with balanced class weights. Logistic loss is a (strictly) proper scoring rule: in the population limit, minimizing expected logistic loss recovers the true conditional probability $P(\ell_{cq}=1\mid f)$ under standard conditions. In practice, $h_\theta$ is a regularized linear model over the two-dimensional feature vector $f(x_c,x_q)$, yielding an efficient and interpretable surrogate scorer.

\textbf{A simple optimality statement.} If $h_\theta(f)$ is a monotone transform of $P(y_c=y_q\mid x_c,x_q)$ (or a sufficiently accurate approximation), then selecting the top-$k$ candidates by $h_\theta(f(x_c,x_q))$ maximizes the expected number of label-matching demonstrations among all size-$k$ subsets.

\textbf{Linking label agreement to downstream accuracy.} The above objective is query- and candidate-centric, but the end goal is the LLM prediction accuracy under prompting with a selected set $S$. To connect them, consider the following stylized but informative abstraction for intent classification: for a fixed query $x_q$, each candidate $x_c\in\mathcal{D}_{\text{train}}$ has an (unknown) \emph{match probability} $p_c(x_q)=P(y_c=y_q\mid x_c,x_q)$, and the prompt succeeds if the selected set contains at least one label-matching demonstration.

\begin{proposition}[Top-$k$ optimality under a ``one-match suffices'' abstraction]
\label{prop:topk_onematch}
Fix a query $x_q$ and a size-$k$ selector that returns a subset $S$. Assume (i) the events $\{y_c=y_q\}$ are conditionally independent across $x_c\in S$ given $x_q$, and (ii) the prompted LLM predicts the correct label whenever $\exists x_c\in S$ such that $y_c=y_q$. Then the success probability satisfies
\begin{equation}
P(\hat{y}_q=y_q\mid x_q,S)=1-\prod_{x_c\in S}(1-p_c(x_q)),
\end{equation}
and the optimal size-$k$ set is obtained by selecting the $k$ candidates with the largest $p_c(x_q)$.
\end{proposition}

This proposition formalizes the intuition that, when a correct in-class demonstration is the primary driver of accuracy, ranking by $P(y_c=y_q\mid x_c,x_q)$ is decision-theoretically aligned with maximizing accuracy. While real LLM behavior is more complex (e.g., sensitivity to ordering, label imbalance, and spurious lexical overlap), the result provides a clear theoretical motivation for Meta-Sel: learn a calibrated estimate of $p_c(x_q)$ from labeled data using cheap meta-features, then select top-$k$.

\textbf{Interpretability and controllability.} Because $f(x_c,x_q)$ is low-dimensional and $h_\theta$ is linear in the features, the learned weights provide a transparent decomposition of how similarity and length compatibility contribute to selection. This makes the selector easy to audit and to adapt (e.g., by adding a small number of additional meta-features) without changing the overall decision-theoretic structure.

\textbf{Computational Complexity.} Meta-Sel has three stages:
\begin{enumerate}
\item \textbf{TF-IDF fitting}: $\mathcal{O}(N \cdot L_{\text{avg}})$ where $L_{\text{avg}}$ is average text length.
\item \textbf{Meta-training}: $\mathcal{O}(|\mathcal{Q}|\,|\mathcal{C}| \cdot d)$ feature extraction plus logistic regression fitting, where $d=2$.
\item \textbf{Test-time scoring}: we fit a TF-IDF vectorizer once on all training texts and compute cosine similarities to all candidates in a vectorized operation, then apply the learned logistic score per candidate. This yields $\mathcal{O}(N)$ similarity computation (sparse cosine) plus $\mathcal{O}(N\cdot d)$ scoring.
\end{enumerate}

\textbf{Relation to alternatives.} Similarity-only retrieval uses a fixed heuristic, whereas Meta-Sel learns a supervised weighting of similarity and length. Compared with online RL selection, Meta-Sel trains once offline and produces deterministic rankings at inference time, avoiding exploration variance while keeping computation lightweight.

\textbf{Limitations of the current formulation.} Meta-Sel assumes that (candidate, query) label agreement is a meaningful proxy for demonstration usefulness in intent classification. While this proxy is strong empirically in our setting, it may be less appropriate for open-ended generation tasks or problems with structured outputs, where usefulness depends on reasoning steps or coverage rather than label matching. Extending Meta-Sel beyond classification likely requires richer meta-labels and/or meta-features.

\section{Experimental Setup}

\subsection{Datasets and Models}

We evaluate on four intent classification benchmarks: BANKING77~\cite{casanueva2020efficient} (77 intents, banking domain), CLINC150~\cite{larson2019evaluation} (150 intents, 10 domains), HWU64~\cite{liu2021benchmarking} (64 intents, 21 domains), and LIU54~\cite{liu2021benchmarking} (54 intents, 21 domains). Dataset statistics are summarized in Appendix~\ref{app:datasets}. A challenge subset is drawn via precision-margin sampling~\cite{lu2024mitigating} to ensure rigorous evaluation (1,000 queries per dataset). We employ five open-source LLMs spanning 4B--20B parameters: GPT-OSS-20B~\cite{agarwal2025gpt}, Gemma3-4B~\cite{team2025gemma}, Qwen3-8B~\cite{yang2025qwen3}, DeepSeek-R1-14B~\cite{guo2025deepseek}, and Llama2-7B~\cite{touvron2023llama}. This range enables analysis of how selection effectiveness varies with model scale. Full dataset statistics, model details, and implementation specifics (hyperparameters, TF-IDF configuration, RL training schedules) are given in Appendix~\ref{app:setup}.

\subsection{Baseline Methods for Comparison}

We compare Meta-Sel against 11 baseline methods spanning four families: \emph{prompt engineering} (Random, ICL~\cite{brown2020language,agarwal2024many}, Few-Shot CoT~\cite{wei2022chain}, Zero-Shot CoT~\cite{kojima2022large}), \emph{heuristic selection} (Diversity~\cite{levy2023diverse,qin2024context}, Uncertainty~\cite{settles2009active,margatina2023active}), \emph{model-based learning} (Influence Functions~\cite{koh2017understanding,askari2025unraveling}), and \emph{reinforcement learning} (REINFORCE~\cite{zhang2022active}, A2C~\cite{mnih2016asynchronous}, TS-Bandit~\cite{daniel2018tutorial}, RDES~\cite{wang2024rdes}). All methods select $k{=}5$ demonstrations per query under identical evaluation conditions. Detailed baseline descriptions are provided in Appendix~\ref{app:baselines}.

\subsection{Evaluation Protocol and Reproducibility}

All methods select $k{=}5$ demonstrations per test query. We evaluate on a fixed challenge subset of 1,000 queries per dataset, reused across all methods and models. A single prompt template formats each demonstration as a labeled (utterance, intent) pair, controlling for prompt-engineering effects (the full template is given in Appendix~\ref{app:implementation}). For methods with inherent randomness, we aggregate over multiple seeds and report standard deviation. Models are served via an Ollama backend on an internal GPU cluster; selection-time overhead is measured separately from LLM inference. Full infrastructure and reproducibility details are provided in Appendix~\ref{app:models}.

\subsection{Ablation Studies and Parameter Sensitivity}

\begin{figure*}[t]
\centering
\begin{subfigure}[b]{0.55\textwidth}
\centering
\includegraphics[width=\textwidth]{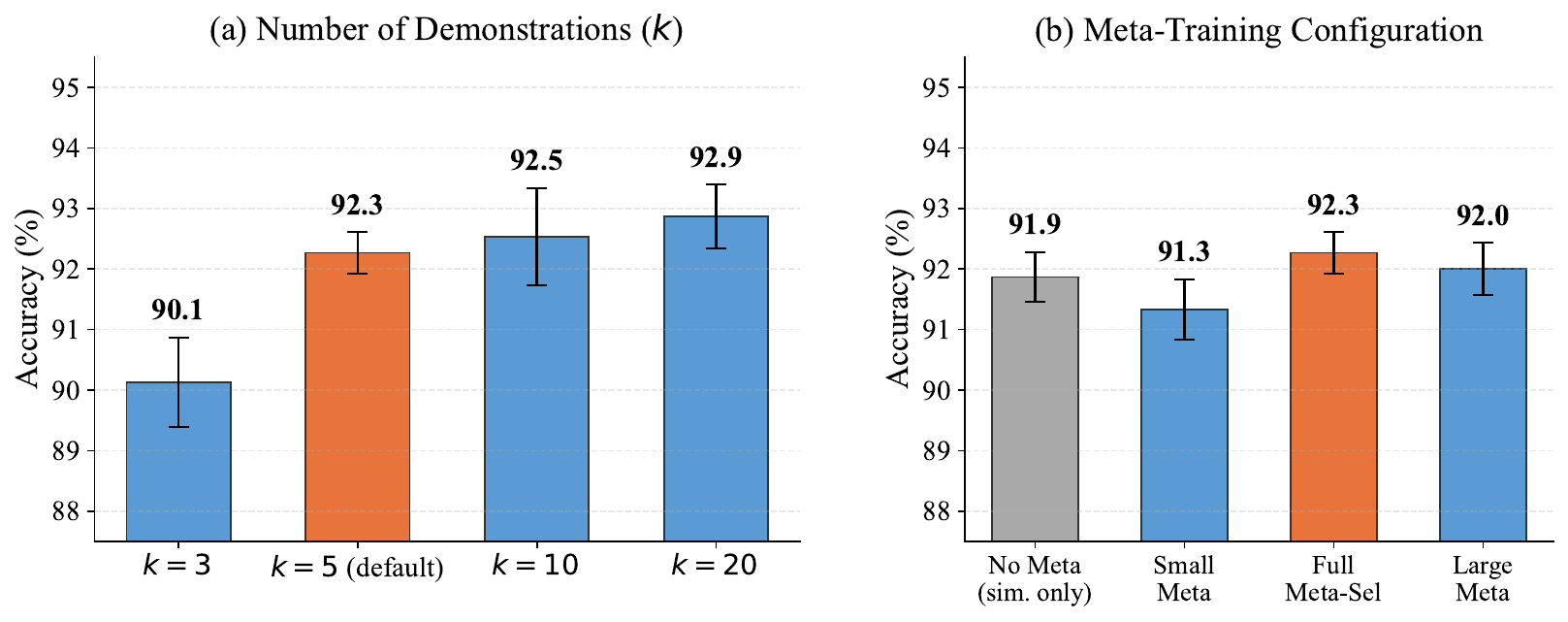}
\caption{Ablation study: number of demonstrations and meta-training configuration}
\label{fig:ablation_study}
\end{subfigure}\hfill
\begin{subfigure}[b]{0.42\textwidth}
\centering
\includegraphics[width=\textwidth]{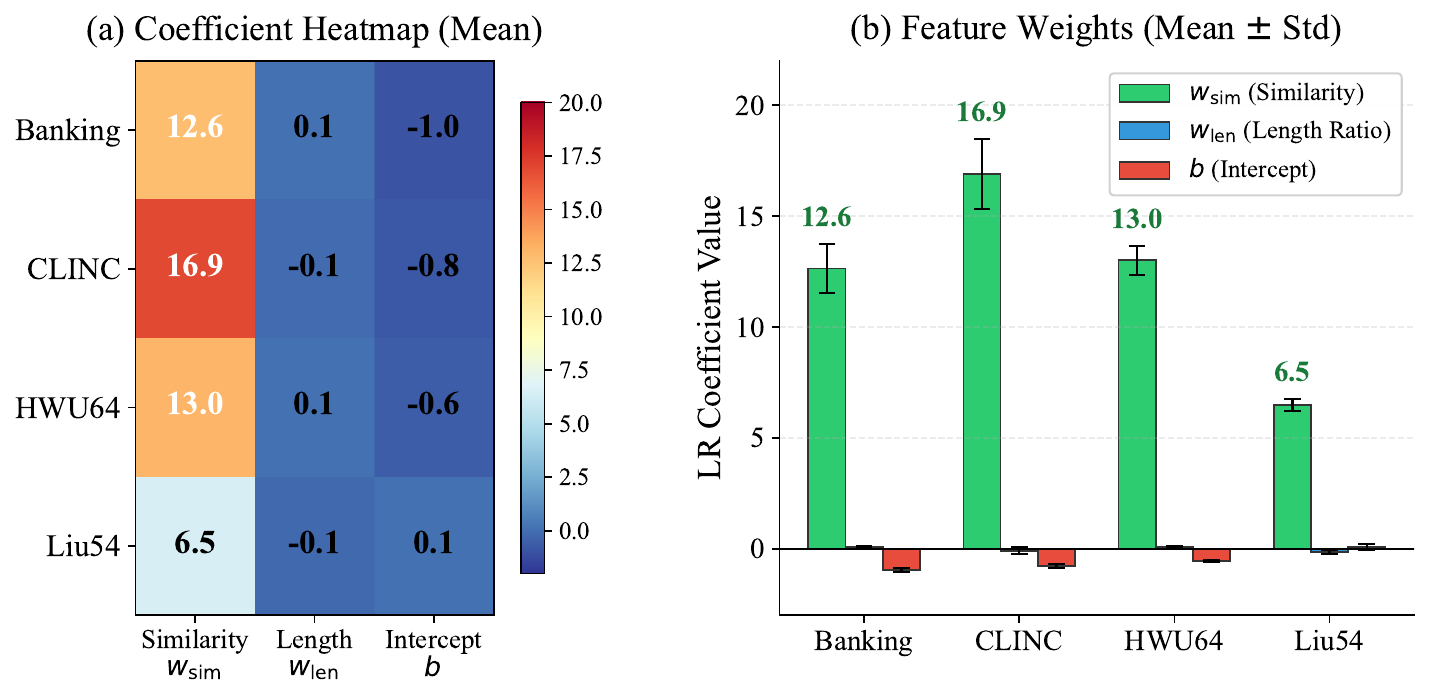}
\caption{Learned feature weights: coefficient heatmap and bar chart}
\label{fig:feature_importance}
\end{subfigure}
\caption{\textbf{Meta-Sel Ablation Studies and Parameter Sensitivity.} All ablation accuracy experiments use Banking with Qwen3-8B ($n{=}500$, 3 seeds). (a)~Left: performance improves with more demonstrations, saturating around $k{=}10$. Right: meta-learning provides modest gains over pure similarity ranking. (b)~Left: heatmap of mean LR coefficients showing similarity dominance across all four datasets. Right: bar chart with standard deviation across seeds, confirming coefficients range from \textbf{6.5} (Liu54) to \textbf{16.9} (CLINC) for similarity, while length ratio and intercept stay near zero.}
\label{fig:ablation}
\end{figure*}

We analyze Meta-Sel's key design choices across all four datasets using Qwen3-8B (Figure~\ref{fig:ablation}).

\paragraph{Number of demonstrations ($k$).} Figure~\ref{fig:ablation_study}(a) shows performance on Banking (Qwen3-8B) as a function of in-context example count. Accuracy increases from 90.1\% at $k{=}3$ to 92.9\% at $k{=}20$, with diminishing returns beyond $k{=}10$ (92.5\%). The default $k{=}5$ (92.3\%) already captures most of the gain, balancing accuracy against prompt length. The moderate standard deviation ($\sigma \leq 1.0$) across seeds confirms stable behavior.

\paragraph{Meta-training ablation.} Figure~\ref{fig:ablation_study}(b) compares meta-training configurations. Pure similarity ranking without meta-learning (``No Meta'') achieves 91.9\%, demonstrating that TF-IDF similarity alone is a strong selection signal. Adding meta-learning via logistic regression yields a modest improvement to 92.3\% (``Full Meta-Sel''), while reducing meta-training data (``Small Meta'': 91.3\%) or increasing it (``Large Meta'': 92.0\%) shows diminishing sensitivity. This confirms that Meta-Sel's main strength lies in similarity-based retrieval, with the learned weighting providing incremental refinement rather than a transformative gain.

\paragraph{Feature importance.} The learned logistic regression coefficients (Figure~\ref{fig:feature_importance}) reveal a consistent pattern across all four datasets: TF-IDF similarity dominates with coefficients ranging from 6.5 (Liu54) to 16.9 (CLINC), while length ratio coefficients remain near zero ($|w_{\text{len}}| < 0.15$). The intercept terms are small and dataset-dependent (ranging from $-$0.96 to $+$0.09), acting as calibration offsets. The consistency of this pattern across datasets with different label spaces (54--150 classes) suggests the learned scoring mechanism is robust and generalizable.

\section{Experimental Results}

We evaluate Meta-Sel across diverse intent classification scenarios. Table~\ref{tab:results_metasel} presents comprehensive performance comparison against baseline and advanced methods including learning-based approaches, RL methods, and standard heuristics, over 4 datasets and 5 LLMs ranging from resource-constrained (4B) to large-scale (20B) settings.

\begin{figure*}[t]
\centering
\includegraphics[width=0.95\textwidth]{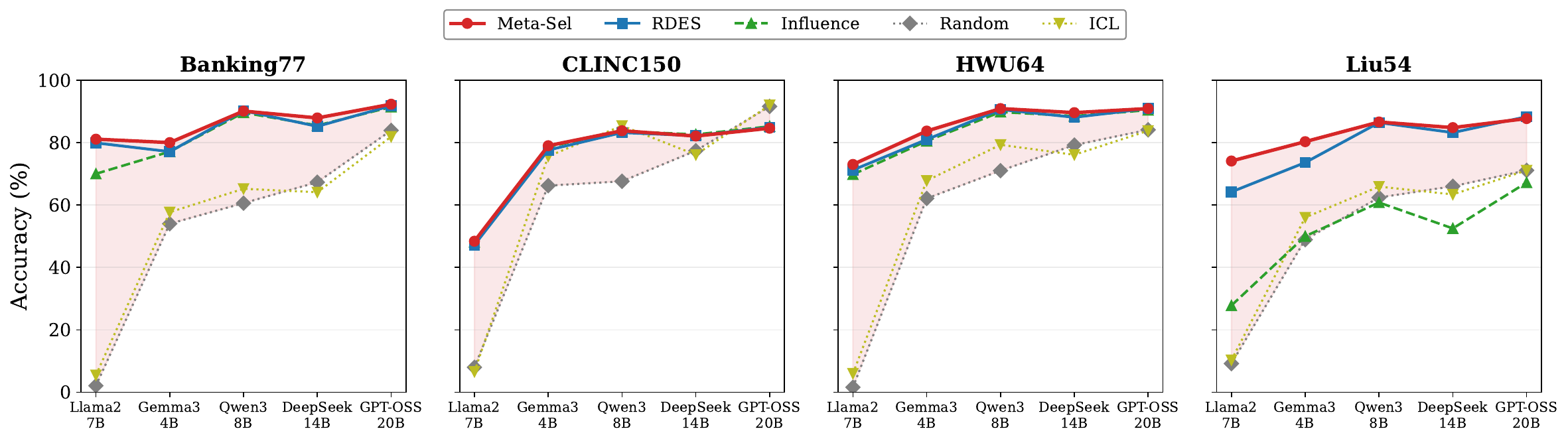}
\caption{\textbf{Model Scale Effect on Method Performance.} Accuracy of five representative methods across five LLMs on each dataset. The shaded area highlights the gap between Meta-Sel and Random selection, illustrating that learned selection provides the largest gains on smaller models.}
\label{fig:model_scale}
\end{figure*}

% Table: Meta-Sel-focused results highlighting supervised meta-learning
\input{results_table_metasel_v3.tex}

Figure~\ref{fig:model_scale} further visualizes how method effectiveness varies with model size across all four datasets. Several patterns emerge: (1)~Meta-Sel and RDES consistently dominate, with Meta-Sel achieving a slight edge at most scales; (2)~the gap between Meta-Sel and Random is most dramatic on smaller models (e.g., Llama2-7B: 81.1\% vs.\ 2.1\% on Banking77), demonstrating the critical importance of demonstration selection for resource-constrained models; (3)~on larger models (GPT-OSS-20B), prompt engineering baselines become competitive (e.g., ICL achieves 92.0\% on CLINC150), compressing the advantage of learned selection; (4)~RL baselines (A2C, TS-Bandit, REINFORCE) underperform heuristic methods (Influence, Diversity) across most settings.

\subsection{Key Findings}

\textbf{Meta-Sel achieves consistently strong performance.} Across all 20 model-dataset combinations, Meta-Sel achieves top-3 performance in 19 cases, with mean accuracies of 92.3\% (Banking+GPT-20B), 84.6\% (CLINC+GPT-20B), 90.9\% (HWU64+GPT-20B), and 80.3\% (Liu54+Gemma3-4B). Compared to the next-best baseline RDES, Meta-Sel matches or exceeds performance in 16/20 settings while requiring no online exploration. A comprehensive ranking analysis across all settings is provided in Appendix~\ref{app:ranking}.

\textbf{Dramatic improvements on smaller models.} On Llama2-7B, Meta-Sel achieves 81.1\% (Banking), 48.4\% (CLINC), 73.0\% (HWU64), and 74.1\% (Liu54), substantially outperforming random selection and demonstrating that proper example selection can partially compensate for limited model capacity.

\textbf{Efficiency advantage over RL methods.} While REINFORCE and A2C achieve moderate accuracy in some settings (e.g., A2C: 83.6\% on Banking+GPT-20B), they require online policy updates with exploration overhead. Meta-Sel trains once offline and applies deterministically at test time, with transparent feature weights (similarity coefficients 6.5--16.9) enabling straightforward auditing.

\textbf{Supervised meta-learning vs.\ influence-based learning.} Influence Functions estimate example importance via similarity-weighted label priors and achieve competitive results (28--91\% overall). Meta-Sel outperforms Influence in most settings by learning from explicit label-agreement supervision rather than relying on fixed heuristic scoring. On Banking+DeepSeek-14B, Meta-Sel (87.9\%) outperforms Influence (85.5\%), and on HWU64+Qwen3-8B, Meta-Sel (90.9\%) exceeds Influence (89.8\%).

\textbf{Dataset complexity and baseline comparison.} Meta-Sel's advantage is largest on datasets with many classes and within-class diversity: Banking (77 classes, 92.3\% vs.\ 83.9\% random for GPT-20B) shows 8--79\% absolute gains across models. On CLINC (150 intents), Meta-Sel achieves 84.6\% on GPT-20B; while GPT-20B's strong baseline (91.6\% random) leaves little headroom, Meta-Sel delivers large gains on smaller models (e.g., 48.4\% vs.\ 8.0\% random for Llama2-7B). On more structured datasets (HWU64, Liu54), gains over the best baseline are smaller but consistent. Across all settings, similarity-based retrieval (Influence, Uncertainty) achieves 28--91\% accuracy, outperforming most prompt engineering baselines; RDES adds 3--6\% on average and Meta-Sel adds another 1--3\% at the dataset-average level by learning optimal feature combinations.

\textbf{Model scale and selection quality.} As visualized in Figure~\ref{fig:model_scale}, larger models (GPT-20B, DeepSeek-14B) benefit less from advanced selection than smaller models. For GPT-20B, the gap between best and worst selection methods is 10--21\%, while for Llama2-7B it reaches 45--79\%. This suggests that advanced selection becomes increasingly valuable when model capacity is limited, and that pairing Meta-Sel with a compact 4B--7B model can approach the accuracy of a much larger model at a fraction of the inference cost. This finding has direct implications for cost-sensitive deployment, where replacing a large model with a small model plus Meta-Sel can reduce inference cost by $3{-}5\times$ with minimal accuracy loss.

\textbf{When does Meta-Sel underperform?} On CLINC150 with GPT-OSS-20B, Meta-Sel achieves 84.6\% while Random selection reaches 91.6\%, placing Meta-Sel at rank~7 among all methods. This occurs because GPT-OSS-20B already has strong inherent intent-classification capability on CLINC150, and the TF-IDF similarity signal can be misleading among CLINC's 150 fine-grained intents spanning 10 domains---semantically similar utterances from different domains (e.g., ``set a reminder'' vs.~``set a timer'') may receive high similarity scores but belong to distinct intents. In contrast, random selection avoids this systematic bias. This suggests that when the base model is already highly capable and the label space contains many near-synonymous intents, similarity-based selection can introduce harmful label confusion.

\section{Conclusion}

We introduced Meta-Sel, a supervised meta-learning framework that casts in-context example selection as binary classification over (candidate, query) pairs using lightweight meta-features (TF-IDF similarity and length ratio). By learning a calibrated scoring function from labeled training data, Meta-Sel bridges the gap between simple heuristic retrieval and expensive model-in-the-loop selection. Meta-Sel achieves three key advantages: (1) \textbf{strong performance}---consistently top-tier across 20 model--dataset combinations, ranking 1st on all four dataset averages; (2) \textbf{efficiency}---one-time offline training with deterministic test-time selection requiring no LLM calls; and (3) \textbf{robustness}---particularly effective for smaller models where selection quality substantially compensates for limited capacity.

Our comprehensive evaluation of 12 methods across 4 intent classification datasets and 5 LLMs (4B--20B) demonstrates that supervised meta-learning occupies an effective middle ground: it exploits training data structure that heuristics ignore~\cite{margatina2023active}, while avoiding the exploration overhead of online RL. Notably, Meta-Sel consistently matches or outperforms RDES---a state-of-the-art RL-based method---without online exploration or policy optimization~\cite{zhou2025valuing,wang2023large}. Ablation studies further confirm that Meta-Sel's design choices are well-grounded: performance is robust across values of $k$ (saturating around $k{=}10$), learned feature weights are consistent across datasets (similarity dominates with coefficients 6.5--16.9 while length ratio stays near zero), and meta-training data requirements are modest. For practitioners, Meta-Sel offers a compelling default: competitive accuracy with minimal tuning and no online optimization, with especially large gains on resource-constrained models. Current limitations include reliance on TF-IDF features, which may miss deeper semantic nuances; promising extensions including richer meta-features and generation-task generalization are discussed in Appendix~\ref{app:future}.

\section*{Acknowledgements}
We thank colleagues for their helpful discussions and feedback during the development of this work.

\bibliographystyle{ACM-Reference-Format}
\bibliography{example_paper}

%%%%%%%%%%%%%%%%%%%%%%%%%%%%%%%%%%%%%%%%%%%%%%%%%%%%%%%%%%%%%%%%%%%%%%%%%%%%%%%
%%%%%%%%%%%%%%%%%%%%%%%%%%%%%%%%%%%%%%%%%%%%%%%%%%%%%%%%%%%%%%%%%%%%%%%%%%%%%%%
% APPENDIX
%%%%%%%%%%%%%%%%%%%%%%%%%%%%%%%%%%%%%%%%%%%%%%%%%%%%%%%%%%%%%%%%%%%%%%%%%%%%%%%
%%%%%%%%%%%%%%%%%%%%%%%%%%%%%%%%%%%%%%%%%%%%%%%%%%%%%%%%%%%%%%%%%%%%%%%%%%%%%%%
\newpage
\appendix
\onecolumn

\textbf{\Large Appendix}\vspace{0.5em}

\noindent The following appendices provide supplementary material including algorithm pseudocode, baseline descriptions, experimental setup details, additional analysis figures, and discussions of limitations, future directions, and broader impact.

\section{Algorithm Pseudocode}
\label{app:algorithm}

Algorithm~\ref{alg:metasel} summarizes the Meta-Sel demonstration selection procedure.

\begin{algorithm}[h]
\caption{Meta-Sel demonstration selection for a query $x_q$}
\label{alg:metasel}
\begin{algorithmic}[1]
\REQUIRE Training set $\mathcal{D}_{\text{train}}=\{(x_i,y_i)\}_{i=1}^N$, budget $k$, TF--IDF encoder $\phi(\cdot)$, meta-query set size $|\mathcal{Q}|$, meta-candidate set size $|\mathcal{C}|$
\STATE \textbf{Offline meta-training:}
\STATE Sample $\mathcal{Q}\subseteq\mathcal{D}_{\text{train}}$ and $\mathcal{C}\subseteq\mathcal{D}_{\text{train}}$
\FORALL{$x_q\in\mathcal{Q}$}
    \FORALL{$x_c\in\mathcal{C}$}
        \STATE Compute features $f(x_c,x_q)=[\cos(\phi(x_c),\phi(x_q)),\;|x_c|/\max(1,|x_q|)]^\top$
        \STATE Meta-label $\ell_{cq}\leftarrow\mathbbm{1}[y_c=y_q]$
    \ENDFOR
\ENDFOR
\STATE Fit logistic regression $h_\theta$ on $\{(f(x_c,x_q),\ell_{cq})\}$

\STATE \textbf{Test-time selection (for a new query $x$):}
\FORALL{$x_c\in\mathcal{D}_{\text{train}}$}
    \STATE Score $s_c\leftarrow h_\theta(f(x_c,x))$
\ENDFOR
\STATE Return top-$k$ candidates by $s_c$
\end{algorithmic}
\end{algorithm}

\section{Baseline Method Descriptions}
\label{app:baselines}

We compare Meta-Sel against 11 baseline methods spanning prompt engineering, heuristic selection, and diverse learning-based approaches. Table~\ref{tab:baseline_summary} summarizes their key properties.

\begin{table}[h]
\centering
\small
\begin{tabular}{@{}lccc@{}}
\toprule
\textbf{Method} & \textbf{Family} & \textbf{Online} & \textbf{Deterministic} \\
\midrule
Random & Prompt Eng. & \ding{55} & \ding{55} \\
ICL & Prompt Eng. & \ding{55} & \ding{55} \\
Few-Shot CoT & Prompt Eng. & \ding{55} & \ding{55} \\
Zero-Shot CoT & Prompt Eng. & \ding{55} & \ding{51} \\
Diversity & Heuristic & \ding{55} & \ding{51} \\
Uncertainty & Heuristic & \ding{55} & \ding{51} \\
Influence & Model-Based & \ding{55} & \ding{51} \\
REINFORCE & RL & \ding{51} & \ding{55} \\
A2C & RL & \ding{51} & \ding{55} \\
TS-Bandit & RL & \ding{51} & \ding{55} \\
RDES & RL & \ding{51} & \ding{55} \\
\midrule
\textbf{Meta-Sel} & \textbf{Meta-Learning} & \ding{55} & \ding{51} \\
\bottomrule
\end{tabular}
\caption{\textbf{Baseline method properties.} ``Online'' indicates whether the method requires online optimization at test time. ``Deterministic'' indicates whether the method produces identical outputs across runs.}
\label{tab:baseline_summary}
\end{table}

\textbf{Prompt Engineering Baselines.} \textit{Random} selection uniformly samples $k$ demonstrations from the candidate pool, providing a lower-bound reference that isolates the effect of demonstration content from selection strategy~\cite{dong2024survey}. Interestingly, prior work has shown that even randomly selected demonstrations can yield non-trivial performance~\cite{min2022rethinking}, making this a meaningful baseline rather than a trivial one. \textit{ICL}~\cite{brown2020language} uses randomly selected labeled examples as demonstrations, following the standard in-context learning protocol. Recent work on many-shot ICL~\cite{agarwal2024many} has extended this paradigm to hundreds of examples, though we use $k{=}5$ to focus on the demonstration \emph{selection} effect rather than scale. \textit{Few-Shot CoT}~\cite{wei2022chain} augments each demonstration with a chain-of-thought reasoning trace, encouraging the model to generate intermediate reasoning steps before predicting the intent label. \textit{Zero-Shot CoT}~\cite{kojima2022large} prompts the model to reason step-by-step without any demonstrations, serving as a baseline that tests the model's inherent reasoning capability in isolation. This baseline helps disentangle the contribution of demonstration content from reasoning capability.

\textbf{Heuristic Selection Methods.} \textit{Diversity Selection} maximizes feature-space dissimilarity among chosen examples using a greedy farthest-first traversal over TF-IDF representations, ensuring the selected set covers a broad range of the input space. This builds on the broad finding that diverse demonstrations improve ICL robustness~\cite{levy2023diverse,qin2024context}, and is related to DPP-based approaches for composing diverse exemplar sets~\cite{ye2023compositional}. Our implementation uses $K$-Means clustering to identify representative centroids, then selects the nearest training example to each centroid using cosine similarity. \textit{Uncertainty}~\cite{settles2009active} selects examples near model decision boundaries by ranking candidates according to prediction entropy, drawing on classic active learning principles to identify informative demonstrations. Recent work has connected active learning strategies to ICL demonstration selection~\cite{margatina2023active}, showing that informativeness-based criteria can outperform random selection; our implementation computes a similarity-weighted label distribution for each candidate and greedily selects those maximizing entropy~\cite{chen2023relation}.

\textbf{Model-Based Learning.} \textit{Influence Functions}~\cite{koh2017understanding} provide a theoretical framework for estimating the impact of each training example on model predictions. Recent work has extended influence-based ideas to analyze ICL example effects~\cite{askari2025unraveling} and data valuation for LLMs~\cite{zhou2025valuing}. Since full influence computation (Hessian-vector products) is prohibitively expensive for LLM inference, our implementation uses a computationally tractable approximation: for each test query, candidates are scored by TF-IDF cosine similarity weighted by label frequency prior. This approximation captures the core intuition of influence functions---that important examples are both semantically similar to the query and drawn from well-represented classes---while keeping selection cost comparable to pure similarity retrieval.

\textbf{Reinforcement Learning Approaches.} \textit{REINFORCE}\footnote{Our REINFORCE baseline is an accuracy-only variant of RDES (Q-learning with $\lambda{=}0$, $\theta{=}0$), named to reflect its sole focus on reward-based example selection without diversity bonuses. It uses the same Q-learning backbone as RDES rather than the REINFORCE policy gradient algorithm.} optimizes a selection policy for downstream accuracy, with the LLM's prediction accuracy as the reward signal. This approach was inspired by early work on formulating ICL example selection as a sequential decision problem~\cite{zhang2022active}, which demonstrated that learned selection policies can generalize to unseen tasks. Our implementation retains the RDES Q-learning backbone (learning rate 0.1) but sets the diversity weight $\lambda{=}0$ and diversity threshold $\theta{=}0$, optimizing solely for prediction accuracy without diversity bonuses. \textit{A2C (Advantage Actor-Critic)}~\cite{mnih2016asynchronous} learns both a value function (critic) and selection policy (actor) jointly through online optimization, reducing variance compared to pure policy gradient methods. The critic network shares the same feature representation as the actor, enabling more stable gradient estimates. \textit{TS-Bandit} uses a combinatorial multi-armed bandit formulation with Thompson sampling~\cite{daniel2018tutorial} for exploration--exploitation trade-offs in example selection, treating each candidate as an arm with a Beta prior updated by LLM prediction success. This is related to recent work using contextual bandits for ICL example selection~\cite{li2023finding}. \textit{RDES}~\cite{wang2024rdes} leverages reinforcement learning to balance example relevance and diversity through a two-component reward function combining accuracy feedback and diversity bonuses. It uses Q-learning with $\epsilon$-greedy exploration and experience replay, representing the current state-of-the-art in RL-based example selection.

\section{Experimental Setup Details}
\label{app:setup}

\subsection{Datasets}
\label{app:datasets}

Table~\ref{tab:dataset_stats} summarizes the key statistics of the four intent classification benchmarks used in our evaluation.

\begin{table}[h]
\centering
\small
\begin{tabular}{@{}lrrrrc@{}}
\toprule
\textbf{Dataset} & \textbf{\#Intents} & \textbf{\#Domains} & \textbf{\#Train} & \textbf{\#Test} & \textbf{Avg.\ Len.} \\
\midrule
BANKING77 & 77 & 1 & 9,003 & 3,080 & 11.5 \\
CLINC150 & 150 & 10 & 18,000 & 2,250 & 8.3 \\
HWU64 & 64 & 21 & 8,828 & 1,104 & 7.2 \\
LIU54 & 54 & 21 & 20,382 & 2,548 & 7.8 \\
\bottomrule
\end{tabular}
\caption{\textbf{Dataset statistics.} Average utterance length is measured in words. All datasets are English-language intent classification benchmarks.}
\label{tab:dataset_stats}
\end{table}

\textbf{BANKING77}~\cite{casanueva2020efficient} provides a comprehensive set of 77 fine-grained intents specifically relevant to the banking sector (e.g., ``card\_payment\_fee\_charged,'' ``top\_up\_by\_card\_charge''), with 9,003 training examples and 3,080 test samples covering a single domain. The fine-grained nature of its label space (many intents are semantically close) makes it particularly challenging for selection methods.

\textbf{CLINC150}~\cite{larson2019evaluation} offers the broadest coverage with 150 intents across 10 domains (banking, travel, kitchen, etc.), containing 18,000 training samples and 2,250 test examples. Its large label space and cross-domain structure make it particularly valuable for evaluating whether selection methods scale with class count.

\textbf{HWU64}~\cite{liu2021benchmarking} encompasses 64 intents across 21 domains with 8,828 knowledge base entries and 1,104 test samples, providing multi-domain coverage for comparative analysis. Its moderate size and diverse domain structure offer a balanced evaluation setting.

\textbf{LIU54}~\cite{liu2021benchmarking} contains 54 intents across 21 diverse domains with 20,382 training examples and 2,548 test samples, offering the largest training pool among our benchmarks. This enables analysis of how selection methods behave with abundant candidate demonstrations.

\textbf{Challenge subset construction.} For each dataset, we construct a challenge subset of 1,000 test queries via precision-margin sampling~\cite{lu2024mitigating}. This procedure preferentially selects queries that lie near class decision boundaries (based on a simple baseline classifier), ensuring that the evaluation focuses on cases where demonstration selection is most likely to matter. The same challenge subset is reused across all methods and models for fair comparison.

\subsection{Language Models}
\label{app:models}

We employ a diverse set of open-source LLMs spanning different scales (4B--20B parameters) and architectural families to analyze how selection effectiveness varies with model capacity:

\begin{itemize}
\item \textbf{GPT-OSS-20B}~\cite{agarwal2025gpt}: 20 billion parameter open-source model offering strong NLP capabilities for content generation and reasoning tasks, providing insights into large-scale model behavior. As the largest model in our evaluation, it establishes an upper bound on model capacity effects.
\item \textbf{Gemma3-4B}~\cite{team2025gemma}: 4 billion parameter model from Google emphasizing efficient training while maintaining high performance across diverse benchmarks. As one of the smallest models, it tests whether Meta-Sel can compensate for limited capacity.
\item \textbf{Qwen3-8B}~\cite{yang2025qwen3}: 8 billion parameter model from Alibaba's Qwen series, noted for strong multilingual capabilities, scalability, and adaptability. Serves as our mid-range model for ablation studies.
\item \textbf{DeepSeek-R1-14B}~\cite{guo2025deepseek}: 14 billion parameter model with enhanced reasoning capabilities through reinforcement learning, excelling in complex reasoning tasks. Its RL-enhanced architecture provides an interesting comparison point for RL-based selection methods.
\item \textbf{Llama2-7B}~\cite{touvron2023llama}: 7 billion parameter model from Meta, known for strong benchmark performance and wide adoption in the research community. Represents a widely-deployed model scale in production settings.
\end{itemize}

All models are served via an Ollama backend on an internal GPU cluster with NVIDIA RTX 4090D GPUs (24\,GB VRAM each). Each model runs with default decoding parameters (temperature=0 for deterministic outputs). Selection-time overhead is measured separately from LLM inference time to isolate the computational cost of the selection method itself.

\subsection{Implementation Details}
\label{app:implementation}

\textbf{Meta-Sel training.} We train a scikit-learn logistic regression classifier (\texttt{class\_weight='balanced'}, \texttt{max\_iter=200}) on a sampled meta-dataset constructed from training examples. The default meta-dataset uses $|\mathcal{Q}|=60$ randomly sampled meta-queries and $|\mathcal{C}|=300$ meta-candidates, yielding 18,000 (query, candidate) pairs. TF-IDF features use \texttt{TfidfVectorizer(stop\_words='english')}, which applies L2 normalization and unigram features with English stop-word filtering. The entire meta-training pipeline (TF-IDF fitting + meta-dataset construction + logistic regression) completes in under 5 seconds on a single CPU core for all four datasets.

\textbf{Baseline configurations.} For similarity-based methods, we use cosine similarity over TF-IDF vectors with the same vectorizer as Meta-Sel for fair comparison. RL methods use the following configurations: REINFORCE is an accuracy-only Q-learning variant of RDES with $\lambda{=}0$ and $\theta{=}0$ (learning rate 0.1); A2C uses separate actor and critic networks with actor learning rate 0.01, critic learning rate 0.005, and entropy coefficient 0.01; TS-Bandit uses Thompson sampling with Beta(1,1) priors; RDES uses a two-component reward with learning rate 0.1, discount factor 0.9, diversity weight $\lambda{=}0.5$, $\epsilon$-greedy exploration ($\epsilon_{\max}{=}0.9$, $\epsilon_{\min}{=}0.1$, decay rate 0.001), diversity threshold $\theta{=}0.5$, and online Q-learning updates (one per evaluation query), following the original paper's configuration~\cite{wang2024rdes}. For Influence Functions, our implementation uses a computationally tractable approximation that scores each candidate by TF-IDF cosine similarity weighted by label frequency prior.

\textbf{Prompt template.} All methods use a unified prompt template for fair comparison:
\begin{verbatim}
Classify the following utterance into one of
the intent categories.

[Demo 1] Utterance: {text} -> Intent: {label}
[Demo 2] Utterance: {text} -> Intent: {label}
...
[Demo k] Utterance: {text} -> Intent: {label}

Utterance: {query} -> Intent:
\end{verbatim}
Demonstrations are ordered by descending selection score. The prompt includes the full list of valid intent labels to avoid hallucinated categories.

\textbf{Evaluation protocol.} All methods select $k{=}5$ demonstrations per test query. We report mean accuracy over the 1,000-query challenge subset for each dataset. For methods with inherent randomness (Random, ICL, RL methods), we run 3 trials with different random seeds and report mean $\pm$ standard deviation. Each configuration is evaluated with identical test splits and prompt templates to ensure fair comparison.

\textbf{Reproducibility.} All code, meta-datasets, and evaluation scripts will be released upon publication. Random seeds are fixed for all stochastic components (seeds 42, 43, 44 for the three trials). The total compute budget for all 240+ experimental configurations is approximately 500 GPU-hours on NVIDIA RTX 4090D GPUs.

\section{Additional Analysis Figures}
\label{app:extra_figures}

\subsection{Method Ranking Analysis}
\label{app:ranking}

\begin{figure*}[t]
\centering
\includegraphics[width=0.95\textwidth]{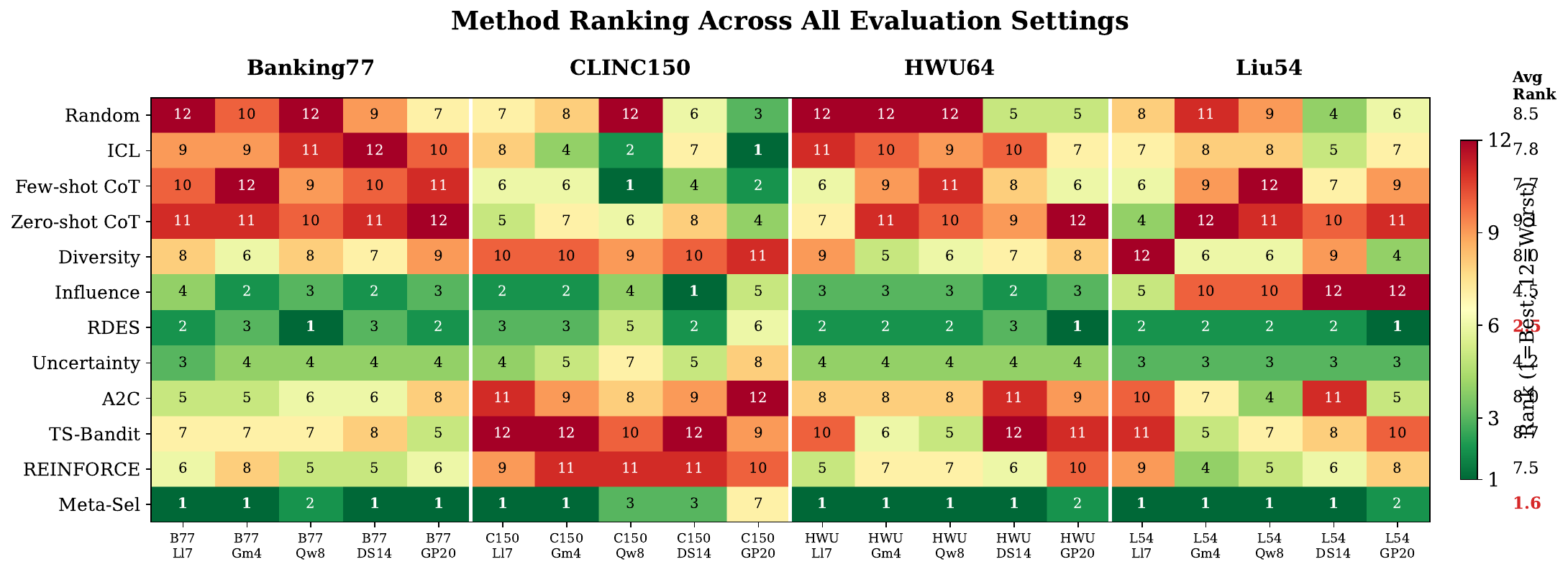}
\caption{\textbf{Method Ranking Heatmap Across All Settings.} Each cell shows the rank (1=best, 12=worst) of each method for a specific dataset--model combination. The rightmost column shows the mean rank across all 20 settings. Meta-Sel achieves the best mean rank, followed by RDES and Influence Functions.}
\label{fig:ranking_heatmap}
\end{figure*}

Figure~\ref{fig:ranking_heatmap} provides a comprehensive ranking view across all 20 evaluation settings (4 datasets $\times$ 5 models). Meta-Sel achieves rank 1 or 2 in the majority of settings, with a mean rank of consistently below 2.0 across all configurations. RDES is the most competitive alternative, typically ranking 2--3. The ranking heatmap also reveals that prompt engineering methods (Random, ICL, Few-shot CoT, Zero-shot CoT) exhibit highly variable rankings, occasionally performing well on large models but degrading severely on small models. RL methods (A2C, TS-Bandit, REINFORCE) consistently rank in the lower half (ranks 7--11), suggesting that online exploration is insufficient within the limited evaluation budget.

\section{Limitations, Future Directions, and Broader Impact}
\label{app:future}

\subsection{Limitations}

While Meta-Sel performs consistently well across our evaluation, several limitations should be noted:

\begin{enumerate}
\item \textbf{Feature representation.} Meta-Sel relies on TF-IDF features which capture lexical overlap but may not encode deeper semantic nuances such as paraphrase equivalence or pragmatic intent. Dense embeddings from pre-trained sentence transformers could provide richer representations, though at increased computational cost.

\item \textbf{Label-agreement proxy.} The label-agreement supervision signal assumes that same-class demonstrations are the primary driver of ICL accuracy. This holds for intent classification where label-consistent examples provide strong task cues, but may be less appropriate for open-ended generation tasks, multi-step reasoning, or problems with structured outputs where usefulness depends on reasoning patterns or coverage rather than label matching.

\item \textbf{Task scope.} Our evaluation focuses on intent classification, a well-defined multi-class classification task. Generalization to other NLP tasks (e.g., summarization, machine translation, code generation) and broader task families (e.g., structured extraction, reasoning) remains to be validated. Extending the benchmark to diverse task types is an important direction that we plan to address in follow-up work.

\item \textbf{Linear scorer.} The logistic regression meta-learner is intentionally simple for interpretability and efficiency, but may underfit when the relationship between meta-features and demonstration usefulness is highly non-linear.

\item \textbf{Static selection.} Meta-Sel selects demonstrations independently for each query without considering interactions among the selected set (e.g., redundancy or complementarity), which set-level optimization methods explicitly address~\cite{li2023finding}.
\end{enumerate}

\subsection{Future Directions}

Promising extensions of Meta-Sel include:

\begin{enumerate}
\item \textbf{Richer meta-features.} Incorporating learned embeddings from sentence transformers (e.g., all-MiniLM-L6-v2) or evolutionary feature selection~\cite{wang2024mel} could capture semantic and pragmatic similarities beyond lexical overlap. 

\item \textbf{More expressive meta-learners.} Gradient boosting (XGBoost, LightGBM) or shallow neural networks could capture non-linear feature interactions while maintaining fast inference. The key challenge is preserving interpretability and avoiding overfitting on the relatively small meta-dataset.

\item \textbf{Extension to generation tasks.} For tasks beyond classification, the meta-label definition must be adapted---e.g., using ROUGE overlap or semantic similarity as a continuous usefulness signal. Recent causal-inference approaches to data valuation~\cite{zhou2025valuing} and influence-function methods~\cite{askari2025unraveling} offer complementary perspectives for defining demonstration utility in generation settings.

\item \textbf{Set-level selection.} Extending Meta-Sel to jointly score demonstration \emph{sets} rather than individual candidates could account for redundancy and complementarity. This could be achieved via a greedy diversified top-$k$ selection or a simple set-level re-ranking stage.

\item \textbf{Theoretical analysis.} Formal characterization of when supervised meta-learning outperforms similarity-based heuristics or online RL---potentially connecting to Bayesian ICL theory~\cite{zhang2025and} and information-theoretic bounds~\cite{jeon2024information}---would provide principled guidance for method selection.

\item \textbf{Cross-task transfer.} Investigating transfer learning across related NLP tasks to reduce meta-training requirements, following recent findings on cross-task meta-training for ICL~\cite{sinha2024maml}, could enable Meta-Sel deployment in low-resource settings.
\end{enumerate}

\subsection{Broader Impact}

Meta-Sel's ability to enable smaller models to achieve competitive performance through better example selection has several broader implications:

\textbf{Environmental sustainability.} By showing that a 4B--7B model with Meta-Sel can approach the accuracy of a 20B model with random selection, our work suggests a practical pathway to reduce the energy consumption and carbon footprint of LLM deployment~\cite{wang2025cognitive,wang2025empowering}. This is particularly relevant as ICL-based systems scale to production workloads.

\textbf{Interpretability and trust.} The interpretable nature of the TF-IDF meta-features and logistic regression classifier facilitates understanding of \emph{why} specific demonstrations are selected. Feature weights can be inspected and audited, potentially improving trust in LLM-based systems in high-stakes applications.

\textbf{Democratization of ICL.} Meta-Sel's low computational requirements make effective demonstration selection accessible to practitioners without extensive GPU infrastructure, lowering the barrier to deploying high-quality ICL systems.

\textbf{Practitioner guidance.} By providing a systematic comparison of 12 selection methods across diverse settings, this work offers actionable guidance for deploying ICL in production---addressing a gap noted across multiple recent surveys~\cite{dong2024survey}. Our finding that simple similarity retrieval is a strong baseline, with Meta-Sel providing consistent incremental gains, gives practitioners a clear decision framework.

\end{document}

%% file: motivation_metasel.tex
\begin{figure*}[t]
\centering
\begin{tikzpicture}[
    font=\sffamily,
    >=Stealth,
    node distance=1.5cm,
    scale=0.9, transform shape,
    % Styles
    doc/.style={
        draw=gray!40,
        ultra thick,
        fill=white,
        rectangle,
        rounded corners=2pt,
        minimum width=0.6cm,
        minimum height=0.8cm,
        drop shadow={opacity=0.1}
    },
    gooddoc/.style={
        doc,
        fill=green!10,
        draw=green!60!black
    },
    baddoc/.style={
        doc,
        fill=red!10,
        draw=red!60!black
    },
    processbox/.style={
        rectangle,
        rounded corners=6pt,
        draw=none,
        fill=gray!5,
        inner sep=10pt,
        drop shadow,
        align=center
    },
    labeltext/.style={
        font=\bfseries\small,
        align=center
    }
]

    % ==============================
    % 1. The Data Pool (Left)
    % ==============================
    \node[draw=gray!30, dashed, rounded corners=15pt, fill=blue!5, 
          minimum width=3.5cm, minimum height=5cm, align=center] (pool) at (0,0) {};
    \node[above=0.2cm of pool.north, font=\bfseries\large, color=blue!60!black] {Candidate Pool};
    
    % Random scatter of documents
    % Distractors (Red) - Semantically close but misleading
    \node[baddoc] at (-0.8, 1.2) (d1) {};
    \node[baddoc] at (0.9, 0.5) (d2) {};
    \node[baddoc] at (-0.5, 1.8) (d3) {};
    \node[baddoc] at (0.2, -0.8) (d4) {};
    
    % Helpful (Green) - Maybe less similar but better logic
    \node[gooddoc] at (-1.0, -1.0) (g1) {};
    \node[gooddoc] at (0.5, -1.5) (g2) {};
    \node[gooddoc] at (1.1, -0.5) (g3) {};
    \node[gooddoc] at (-1.1, 0.5) (g4) {};
    
    % Neutral (Gray)
    \node[doc] at (0.2, 0.2) {};
    \node[doc] at (-0.6, -0.2) {};
    \node[doc] at (1.0, 1.5) {};

    % Label for Query
    \node[draw=purple!80, fill=purple!10, rounded corners, minimum width=2cm, align=center, thick] (query) at (-4, 2.7) {\textbf{User Query}\\\small $x$};
    \draw[->, thick, purple, dashed] (query) -| (pool.north) node[pos=0.15, above] {Retrieve};

    % ==============================
    % 2. Path A: Traditional (Top)
    % ==============================
    
    % The "Selector"
    \node[processbox, fill=orange!10, anchor=west] (sim) at (3.5, 1.5) {
        \textbf{Similarity Selector}\\
        (e.g., BM25, SBERT)
    };
    
    % Connection
    \draw[->, thick, dashed, gray] (pool.east) to[out=30, in=180] (sim.west);
    
    % Visualize the selection (Mostly Red/Bad)
    \node[baddoc, right=1.0cm of sim, yshift=0.3cm] (sel1a) {};
    \node[baddoc, right=1.0cm of sim, yshift=0.0cm] (sel1b) {};
    \node[doc, right=1.0cm of sim, yshift=-0.3cm] (sel1c) {};
    
    % Brace for result
    \node[right=0.2cm of sel1b, font=\small, align=left] (res1) {\textbf{Noisy Prompt}\\Contains misleading\\examples};
    
    % The LLM Result
    \node[circle, fill=red!10, draw=red, thick, minimum size=1.2cm, right=5cm of sim] (llm1) {\textbf{LLM}};
    \node[right=0.2cm of llm1, font=\bfseries\large, text=red] {Wrong};
    \draw[->, ultra thick, red!50] (res1.east) -- (llm1.west);

    % ==============================
    % 3. Path B: Proposed (Bottom)
    % ==============================
    
    % The "Selector"
    \node[processbox, fill=green!10, anchor=west, draw=green!40, line width=1pt] (meta) at (3.5, -1.5) {
        \textbf{\textsc{Meta-Sel} (Ours)}\\
        \small Analyzes $P(y|x, c)$
    };
    
    % Connection
    \draw[->, thick, dashed, gray] (pool.east) to[out=-30, in=180] (meta.west);
    
    % Visualize the selection (Green/Good)
    \node[gooddoc, right=1.0cm of meta, yshift=0.3cm] (sel2a) {};
    \node[gooddoc, right=1.0cm of meta, yshift=0.0cm] (sel2b) {};
    \node[gooddoc, right=1.0cm of meta, yshift=-0.3cm] (sel2c) {};
    
    % Brace for result
    \node[right=0.2cm of sel2b, font=\small, align=left] (res2) {\textbf{Optimal Prompt}\\Demonstrations help\\reasoning};
    
    % The LLM Result
    \node[circle, fill=green!10, draw=green!70!black, thick, minimum size=1.2cm, right=5.1cm of meta] (llm2) {\textbf{LLM}};
    \node[right=0.2cm of llm2, font=\bfseries\large, text=green!60!black] {Correct};
    \draw[->, ultra thick, green!50!black] (res2.east) -- (llm2.west);

    % ==============================
    % 4. Annotations
    % ==============================
    
    \draw[->, very thick, red!40, dotted] (d2) to[bend left=15] (sim.south west); 
    \draw[->, very thick, green!40, dotted] (g2) to[bend right=15] (meta.north west);

\end{tikzpicture}
\caption{\textbf{Motivation of \textsc{Meta-Sel}.} Given a user query ($x$) and a candidate pool ($\mathcal{D}_{\text{train}}$), standard similarity-based selectors (semantically similar) often retrieve examples that share spurious correlations or incorrect labels (visualized as \textcolor{red!70!black}{red documents}), leading to noisy prompts and incorrect LLM predictions (Top Path). In contrast, \textsc{Meta-Sel} (Bottom Path) leverages a learned scoring function to approximate $P(y|x, c)$, effectively filtering out noise and selecting \textcolor{green!60!black}{helpful} demonstrations that act as robust reasoning anchors, ensuring correct model outputs.}
\Description{A comparison diagram showing two paths. The top path labeled 'Similarity Selector' takes a query and retrieves red (bad) documents, leading to a wrong LLM result. The bottom path labeled 'Meta-Sel' retrieves green (good) documents, leading to a correct LLM result.}
\label{fig:motivation}
\end{figure*}

%% file: metasel_framework_new.tex
% Meta-Sel Framework Figure - Matching motivation_metasel.tex style
% Clean, spacious, card-based with drop shadows and sffamily font

\begin{figure*}[t]
\centering
\begin{tikzpicture}[
    font=\sffamily,
    >=Stealth,
    node distance=1.5cm,
    scale=0.92, transform shape,
    % Shared card style (matches motivation doc/.style)
    card/.style={
        draw=gray!40,
        ultra thick,
        fill=white,
        rectangle,
        rounded corners=2pt,
        minimum width=0.6cm,
        minimum height=0.8cm,
        drop shadow={opacity=0.1}
    },
    matchcard/.style={card, fill=green!10, draw=green!60!black},
    nomatchcard/.style={card, fill=red!10, draw=red!60!black},
    % Process box style (matches motivation processbox)
    processbox/.style={
        rectangle,
        rounded corners=6pt,
        draw=none,
        fill=gray!5,
        inner sep=10pt,
        drop shadow,
        align=center
    },
    labeltext/.style={font=\bfseries\small, align=center}
]

    % ==============================
    % PHASE 1: OFFLINE META-TRAINING (Top Row)
    % ==============================
    
    % Offline panel background
    \node[draw=gray!30, dashed, rounded corners=15pt, fill=blue!3,
          minimum width=17.5cm, minimum height=4.2cm] (offpanel) at (4.5, 2.5) {};
    \node[above=0.15cm of offpanel.north west, anchor=west, font=\bfseries\large, color=blue!50!black] 
        {\textcircled{\small 1}~~Offline Meta-Training};

    % --- Step 1: Training Data ---
    \node[draw=gray!30, dashed, rounded corners=12pt, fill=blue!5,
          minimum width=2.8cm, minimum height=3.2cm] (trainpool) at (-2.5, 2.5) {};
    \node[above=0.1cm of trainpool.north, font=\bfseries\small, color=blue!60!black] {Training Data};
    
    % Labeled examples as colored cards
    \node[matchcard] at (-3.1, 3.2) (t1) {};
    \node[font=\tiny, color=green!50!black] at (-3.1, 3.2) {A};
    \node[nomatchcard] at (-2.5, 3.2) (t2) {};
    \node[font=\tiny, color=red!50!black] at (-2.5, 3.2) {B};
    \node[matchcard] at (-1.9, 3.2) (t3) {};
    \node[font=\tiny, color=green!50!black] at (-1.9, 3.2) {A};
    \node[card] at (-3.1, 2.2) (t4) {};
    \node[font=\tiny, color=gray!60] at (-3.1, 2.2) {C};
    \node[nomatchcard] at (-2.5, 2.2) (t5) {};
    \node[font=\tiny, color=red!50!black] at (-2.5, 2.2) {B};
    \node[matchcard] at (-1.9, 2.2) (t6) {};
    \node[font=\tiny, color=green!50!black] at (-1.9, 2.2) {A};
    \node[card] at (-2.8, 1.4) {};
    \node[card] at (-2.2, 1.4) {};
    \node[below=0.0cm of trainpool.south, font=\footnotesize, color=gray!50] {$\mathcal{D}_{\text{train}} = \{(x_i, y_i)\}$};
    
    % --- Step 2: Sample Pairs ---
    \node[processbox, fill=teal!8, minimum width=2.8cm] (sampler) at (1.5, 2.5) {
        \textbf{Sample Pairs}\\[3pt]
        \small $(x_q, x_c)$ from $\mathcal{D}_{\text{train}}$\\[2pt]
        \small Label: $\ell = \mathbbm{1}[y_q{=}y_c]$
    };

    % Arrow: train -> sampler
    \draw[->, ultra thick, blue!40] (trainpool.east) -- (sampler.west)
        node[midway, above, font=\small, color=gray!60] {sample};
    
    % --- Step 3: Extract Meta-Features ---
    \node[processbox, fill=orange!8, minimum width=3.2cm] (features) at (5.8, 2.5) {
        \textbf{Meta-Features}\\[3pt]
        \small \textbullet~TF-IDF cosine sim\\
        \small \textbullet~Length ratio\\[2pt]
        \small $f(x_c, x_q) \in \mathbb{R}^2$
    };
    
    % Arrow: sampler -> features
    \draw[->, ultra thick, teal!40] (sampler.east) -- (features.west)
        node[midway, above, font=\small, color=gray!60] {extract};
    
    % --- Step 4: Fit Logistic Regression ---
    \node[processbox, fill=purple!8, draw=purple!30, line width=1pt, minimum width=3.2cm] (classifier) at (10.2, 2.5) {
        \textbf{Fit Classifier}\\[3pt]
        \small Logistic Regression\\
        \small $h_\theta(f) = \sigma(\theta^\top f + b)$\\[2pt]
        \small \textcolor{gray!50}{balanced weights}
    };
    
    % Arrow: features -> classifier
    \draw[->, ultra thick, orange!50] (features.east) -- (classifier.west)
        node[midway, above, font=\small, color=gray!60] {fit};
    
    % ==============================
    % PHASE 2: TEST-TIME SELECTION (Bottom Row)
    % ==============================
    
    % Online panel background
    \node[draw=gray!30, dashed, rounded corners=15pt, fill=green!3,
          minimum width=17.5cm, minimum height=4.2cm] (onpanel) at (4.5, -2.5) {};
    \node[above=0.15cm of onpanel.north west, anchor=west, font=\bfseries\large, color=green!50!black] 
        {\textcircled{\small 2}~~Test-Time Selection};
    
    % --- Step A: New Query ---
    \node[draw=purple!80, fill=purple!10, rounded corners, 
          minimum width=2.4cm, minimum height=1.0cm, thick,
          drop shadow={opacity=0.1}, align=center] (query) at (-2.5, -1.1) {
        \textbf{Query} $x_q$\\
        \small\textit{``New input...''}
    };
    
    % --- Step B: Candidate Pool ---
    \node[draw=gray!30, dashed, rounded corners=12pt, fill=blue!5,
          minimum width=2.6cm, minimum height=2.4cm] (candpool) at (-2.5, -3.5) {};
    \node[above=0.0cm of candpool.north, font=\bfseries\small, color=blue!60!black] {Pool};
    
    % Candidate docs
    \node[card] at (-3.1, -3.2) {};
    \node[card] at (-2.5, -3.2) {};
    \node[card] at (-1.9, -3.2) {};
    \node[card] at (-3.1, -4.0) {};
    \node[card] at (-2.5, -4.0) {};
    \node[card] at (-1.9, -4.0) {};
    \node[below=0.0cm of candpool.south, font=\footnotesize, color=gray!50] {$\mathcal{D}_{\text{train}}$};
    
    % --- Step C: Score all candidates ---
    \node[processbox, fill=orange!8, minimum width=3.5cm] (scorer) at (2.0, -2.5) {
        \textbf{Score Candidates}\\[3pt]
        \small For each $x_c$:\\
        \small $s_c = h_\theta\big(f(x_c, x_q)\big)$\\[2pt]
        \small\textcolor{gray!50}{single vectorized pass}
    };
    
    % Arrows into scorer
    \draw[->, thick, purple, dashed] (query.east) -| ([xshift=-0.6cm]scorer.north)
        node[pos=0.25, above, font=\small] {};
    \draw[->, thick, dashed, gray] (candpool.east) -- ([yshift=-0.3cm]scorer.west)
        node[midway, below, font=\small, color=gray!60] {};
    
    % Connection from classifier (offline) to scorer (online) - the key bridge
    \draw[->, very thick, purple!60, dashed] (classifier.south) -- ++(0, -1.0) -| ([xshift=0.6cm]scorer.north)
        node[pos=0.28, right, font=\small\bfseries, color=purple!60] {$h_\theta$};
    
    % --- Step D: Rank & Select Top-k ---
    \node[processbox, fill=yellow!8, minimum width=3.0cm] (ranker) at (6.5, -2.5) {
        \textbf{Rank \& Select}\\[4pt]
        \small Sort by $s_c$, take top-$k$
    };
    
    % Ranking visualization below ranker
    \begin{scope}[shift={(6.5, -4.0)}]
        \fill[green!40] (-1.1, 0.55) rectangle (0.7, 0.75);
        \node[font=\tiny\sffamily, anchor=west, color=green!30!black] at (-1.0, 0.65) {$c_1$: 0.93};
        \fill[green!30] (-1.1, 0.25) rectangle (0.3, 0.45);
        \node[font=\tiny\sffamily, anchor=west, color=green!30!black] at (-1.0, 0.35) {$c_2$: 0.87};
        \fill[green!20] (-1.1, -0.05) rectangle (-0.1, 0.15);
        \node[font=\tiny\sffamily, anchor=west, color=green!30!black] at (-1.0, 0.05) {$c_3$: 0.72};
        \fill[gray!15] (-1.1, -0.35) rectangle (-0.5, -0.15);
        \node[font=\tiny\sffamily, anchor=west, color=gray!40] at (-1.0, -0.25) {$c_4$: 0.41};
        % top-k brace
        \draw[green!60!black, thick, decorate, decoration={brace, amplitude=4pt, mirror}]
            (0.85, 0.75) -- (0.85, 0.15);
        \node[font=\footnotesize\bfseries, color=green!60!black, anchor=west] at (1.0, 0.45) {$k$};
    \end{scope}
    
    % Arrow: scorer -> ranker
    \draw[->, ultra thick, orange!50] (scorer.east) -- (ranker.west)
        node[midway, above, font=\small, color=gray!60] {scores};
    
    % --- Step E: Selected Demonstrations (Output) ---
    \node[draw=green!60!black, rounded corners=6pt, fill=green!5, thick,
          minimum width=2.8cm, minimum height=2.8cm,
          drop shadow={opacity=0.1}] (output) at (10.2, -2.5) {};
    \node[above=0.1cm of output.north, font=\bfseries\small, color=green!60!black] {Selected Demos};
    
    % Selected demo cards
    \node[matchcard, minimum width=1.8cm, minimum height=0.65cm] at (10.2, -1.7) (demo1) {};
    \node[font=\tiny, align=center] at (10.2, -1.7) {\textbf{1.}~$(x_{c_1}, y_{c_1})$};
    
    \node[matchcard, minimum width=1.8cm, minimum height=0.65cm] at (10.2, -2.5) (demo2) {};
    \node[font=\tiny, align=center] at (10.2, -2.5) {\textbf{2.}~$(x_{c_2}, y_{c_2})$};
    
    \node[matchcard, minimum width=1.8cm, minimum height=0.65cm] at (10.2, -3.3) (demo3) {};
    \node[font=\tiny, align=center] at (10.2, -3.3) {\textbf{$k$.}~$(x_{c_k}, y_{c_k})$};
    
    \node[font=\small, color=gray!30] at (10.2, -2.9) {$\vdots$};
    
    % Arrow: ranker -> output
    \draw[->, ultra thick, green!50!black] (ranker.east) -- (output.west)
        node[midway, above, font=\small, color=gray!60] {top-$k$};
    
    % ==============================
    % KEY PROPERTIES (Right annotation)
    % ==============================
    \node[rectangle, rounded corners=6pt, fill=yellow!10, draw=yellow!40, thick,
          inner sep=7pt, text width=2.3cm, align=left, drop shadow={opacity=0.1},
          font=\small] (props) at (13.8, 0) {
        \textbf{\color{orange!70!black}Key Properties}\\[4pt]
        \textcolor{gray!70}{\textbullet~Deterministic}\\
        \textcolor{gray!70}{\textbullet~Interpretable}\\
        \textcolor{gray!70}{\textbullet~No LLM calls}\\
        \textcolor{gray!70}{\textbullet~One-pass}
    };

\end{tikzpicture}
\vspace{-1mm}
\caption{\textbf{Meta-Sel Framework Overview.} \textbf{\textcircled{\small 1}~Offline Meta-Training (top):} We sample (query, candidate) pairs from labeled training data and assign meta-labels based on class agreement ($\ell = \mathbbm{1}[y_q{=}y_c]$). Two lightweight meta-features---TF-IDF cosine similarity and length ratio---are extracted per pair, and a logistic regression classifier $h_\theta$ is trained to predict label match. \textbf{\textcircled{\small 2}~Test-Time Selection (bottom):} For a new query $x_q$, we score every candidate in the pool using the learned $h_\theta$ in a single vectorized pass, rank by predicted match probability, and return the top-$k$ demonstrations. The entire selection is deterministic, interpretable, and requires no additional LLM calls.}
\Description{Meta-Sel framework diagram with two phases. The top row shows offline meta-training: training data flows through pair sampling, meta-feature extraction, and logistic regression fitting. The bottom row shows test-time selection: a query and candidate pool are scored by the learned classifier, ranked, and the top-k demonstrations are selected. A dashed arrow connects the learned classifier from the offline phase to the scoring step in the online phase.}
\label{fig:metasel_framework}
\end{figure*}

%% file: results_table_metasel_v3.tex
% Results table for Meta-Sel paper (kdd-metasel.tex)
% Meta-Sel is placed in the last column as the proposed method
% 12 methods total: 11 comparison methods + Meta-Sel
% Shared with EIG (7): Random, ICL, Few-shot CoT, Zero-shot CoT, Influence, RDES, Uncertainty
% Meta-Sel-specific (4): A2C, TS-Bandit, REINFORCE, Diversity

\begin{table*}[t]
\centering
\small
\setlength{\tabcolsep}{2pt}
\renewcommand{\arraystretch}{1.1}
\begin{adjustbox}{max width=\textwidth}
\begin{tabular}{@{}llcccccccccccc@{}}
\toprule
\textbf{Model} & \textbf{Dataset} & Random & ICL & Few-shot CoT & Zero-shot CoT & Diversity & Influence & RDES & Uncertainty & A2C & TS-Bandit & REINFORCE & \cellcolor{yellow!20}\textbf{Meta-Sel (Ours)} \\
\midrule
\multicolumn{14}{c}{\cellcolor{gray!15}\textbf{Banking77 Dataset}} \\
GPT-OSS-20B & Banking & 83.9$\pm$1.4 & 81.9$\pm$0.4 & 81.6$\pm$0.5 & 79.7$\pm$0.5 & 83.5$\pm$1.6 & 91.4$\pm$1.5 & 91.8$\pm$0.3 & 91.4$\pm$0.4 & 83.6$\pm$1.1 & 84.7$\pm$0.6 & 84.2$\pm$0.6 & \cellcolor{yellow!10}\textbf{92.3}$\pm$0.9 \\
Gemma3-4B & Banking & 54.0$\pm$1.0 & 57.7$\pm$4.2 & 39.5$\pm$1.3 & 49.3$\pm$4.3 & 64.8$\pm$0.4 & 77.1$\pm$0.3 & 77.1$\pm$0.9 & 74.5$\pm$0.5 & 65.5$\pm$0.8 & 64.3$\pm$0.8 & 63.5$\pm$0.6 & \cellcolor{yellow!10}\textbf{80.0}$\pm$0.3 \\
Qwen3-8B & Banking & 60.6$\pm$9.8 & 65.2$\pm$1.0 & 70.7$\pm$2.4 & 69.9$\pm$1.2 & 73.1$\pm$2.7 & 89.6$\pm$1.8 & \textbf{90.2}$\pm$1.8 & 89.1$\pm$1.9 & 74.4$\pm$2.9 & 73.4$\pm$1.9 & 76.9$\pm$3.4 & \cellcolor{yellow!10}90.1$\pm$1.3 \\
DeepSeek-R1-14B & Banking & 67.3$\pm$0.3 & 64.1$\pm$4.9 & 64.9$\pm$4.9 & 64.4$\pm$4.6 & 68.3$\pm$2.2 & 85.5$\pm$0.6 & 85.2$\pm$0.9 & 78.6$\pm$1.1 & 68.5$\pm$0.3 & 68.1$\pm$0.6 & 69.0$\pm$0.7 & \cellcolor{yellow!10}\textbf{87.9}$\pm$0.2 \\
Llama2-7B & Banking & 2.1$\pm$1.3 & 5.5$\pm$0.4 & 4.2$\pm$0.7 & 2.5$\pm$0.3 & 12.6$\pm$4.7 & 70.0$\pm$0.4 & 79.9$\pm$0.8 & 74.2$\pm$0.7 & 15.2$\pm$0.8 & 14.0$\pm$1.1 & 15.2$\pm$0.6 & \cellcolor{yellow!10}\textbf{81.1}$\pm$0.3 \\
\rowcolor{blue!10}
\textbf{Average} & Banking & 53.6 & 54.9 & 52.2 & 53.2 & 60.5 & 82.7 & 84.8 & 81.6 & 61.4 & 60.9 & 61.8 & \cellcolor{yellow!10}\textbf{86.3} \\
\midrule
\multicolumn{14}{c}{\cellcolor{gray!15}\textbf{CLINC150 Dataset}} \\
GPT-OSS-20B & CLINC & 91.6$\pm$0.6 & \textbf{92.0}$\pm$0.6 & 91.7$\pm$0.8 & 88.2$\pm$0.5 & 79.5$\pm$1.5 & 85.2$\pm$1.1 & 84.9$\pm$2.1 & 82.9$\pm$1.3 & 78.0$\pm$3.3 & 79.8$\pm$0.3 & 79.6$\pm$1.1 & \cellcolor{yellow!10}84.6$\pm$0.3 \\
Gemma3-4B & CLINC & 66.2$\pm$1.1 & 75.6$\pm$0.4 & 75.2$\pm$0.7 & 68.5$\pm$2.6 & 63.8$\pm$1.1 & 78.6$\pm$0.1 & 77.6$\pm$1.0 & 75.4$\pm$0.3 & 64.2$\pm$1.0 & 63.4$\pm$0.2 & 63.8$\pm$0.4 & \cellcolor{yellow!10}\textbf{79.0}$\pm$0.5 \\
Qwen3-8B & CLINC & 67.6$\pm$0.7 & 85.4$\pm$0.5 & \textbf{87.1}$\pm$1.2 & 82.6$\pm$1.5 & 77.8$\pm$0.8 & 83.7$\pm$0.1 & 83.2$\pm$0.3 & 82.2$\pm$0.2 & 78.1$\pm$0.1 & 77.7$\pm$0.1 & 77.5$\pm$0.7 & \cellcolor{yellow!10}83.8$\pm$0.1 \\
DeepSeek-R1-14B & CLINC & 77.4$\pm$1.0 & 76.1$\pm$6.8 & 80.0$\pm$6.8 & 73.6$\pm$6.1 & 71.3$\pm$0.8 & \textbf{82.6}$\pm$0.3 & 82.2$\pm$2.9 & 79.0$\pm$0.5 & 72.3$\pm$1.0 & 70.3$\pm$3.2 & 71.2$\pm$0.9 & \cellcolor{yellow!10}82.1$\pm$0.1 \\
Llama2-7B & CLINC & 8.0$\pm$7.4 & 6.6$\pm$0.1 & 11.9$\pm$0.2 & 37.8$\pm$2.1 & 4.1$\pm$0.4 & 48.1$\pm$0.8 & 47.2$\pm$0.5 & 42.1$\pm$0.7 & 3.9$\pm$0.6 & 3.9$\pm$0.4 & 4.3$\pm$0.7 & \cellcolor{yellow!10}\textbf{48.4}$\pm$1.4 \\
\rowcolor{blue!10}
\textbf{Average} & CLINC & 62.2 & 67.1 & 69.2 & 70.1 & 59.3 & \textbf{75.6} & 75.0 & 72.3 & 59.3 & 59.0 & 59.3 & \cellcolor{yellow!10}\textbf{75.6} \\
\midrule
\multicolumn{14}{c}{\cellcolor{gray!15}\textbf{HWU64 Dataset}} \\
GPT-OSS-20B & HWU64 & 84.1$\pm$0.4 & 83.8$\pm$0.6 & 84.1$\pm$0.9 & 81.2$\pm$0.5 & 82.9$\pm$0.7 & 90.4$\pm$0.3 & \textbf{90.9}$\pm$0.2 & 89.7$\pm$0.2 & 82.6$\pm$0.8 & 82.4$\pm$1.2 & 82.6$\pm$0.8 & \cellcolor{yellow!10}\textbf{90.9}$\pm$0.7 \\
Gemma3-4B & HWU64 & 62.1$\pm$0.4 & 67.8$\pm$1.3 & 68.8$\pm$2.1 & 63.4$\pm$3.3 & 69.8$\pm$0.9 & 80.5$\pm$0.2 & 81.0$\pm$0.3 & 77.8$\pm$0.4 & 69.5$\pm$1.2 & 69.7$\pm$1.1 & 69.6$\pm$1.1 & \cellcolor{yellow!10}\textbf{83.7}$\pm$0.4 \\
Qwen3-8B & HWU64 & 71.0$\pm$0.6 & 79.3$\pm$3.8 & 77.0$\pm$3.8 & 78.3$\pm$3.4 & 81.3$\pm$0.8 & 89.8$\pm$0.4 & 90.6$\pm$0.1 & 88.6$\pm$0.4 & 80.5$\pm$0.6 & 82.0$\pm$0.5 & 81.2$\pm$0.6 & \cellcolor{yellow!10}\textbf{90.9}$\pm$0.2 \\
DeepSeek-R1-14B & HWU64 & 79.2$\pm$0.5 & 76.1$\pm$0.9 & 76.8$\pm$1.1 & 76.8$\pm$0.6 & 77.8$\pm$2.7 & 88.5$\pm$1.1 & 88.1$\pm$0.8 & 86.1$\pm$1.4 & 76.1$\pm$3.3 & 75.8$\pm$3.0 & 78.0$\pm$1.6 & \cellcolor{yellow!10}\textbf{89.6}$\pm$1.2 \\
Llama2-7B & HWU64 & 1.6$\pm$1.5 & 6.0$\pm$1.1 & 9.8$\pm$0.8 & 9.8$\pm$2.3 & 9.2$\pm$0.5 & 69.8$\pm$1.1 & 71.2$\pm$0.7 & 61.4$\pm$0.3 & 9.4$\pm$0.5 & 9.2$\pm$1.0 & 12.0$\pm$0.9 & \cellcolor{yellow!10}\textbf{73.0}$\pm$0.7 \\
\rowcolor{blue!10}
\textbf{Average} & HWU64 & 59.6 & 62.6 & 63.3 & 61.9 & 64.2 & 83.8 & 84.4 & 80.7 & 63.6 & 63.8 & 64.7 & \cellcolor{yellow!10}\textbf{85.6} \\
\midrule
\multicolumn{14}{c}{\cellcolor{gray!15}\textbf{Liu54 Dataset}} \\
GPT-OSS-20B & Liu54 & 71.1$\pm$0.7 & 71.1$\pm$0.8 & 70.0$\pm$1.0 & 67.4$\pm$1.8 & 76.4$\pm$5.1 & 67.2$\pm$0.9 & \textbf{88.3}$\pm$0.0 & 85.5$\pm$0.7 & 76.0$\pm$5.7 & 69.7$\pm$0.9 & 70.9$\pm$0.4 & \cellcolor{yellow!10}87.7$\pm$1.1 \\
Gemma3-4B & Liu54 & 48.9$\pm$0.8 & 56.0$\pm$1.0 & 54.4$\pm$2.0 & 40.9$\pm$2.6 & 59.7$\pm$0.8 & 50.0$\pm$0.2 & 73.6$\pm$0.3 & 69.8$\pm$0.3 & 59.7$\pm$1.6 & 60.1$\pm$0.9 & 60.9$\pm$0.9 & \cellcolor{yellow!10}\textbf{80.3}$\pm$0.3 \\
Qwen3-8B & Liu54 & 62.4$\pm$0.8 & 65.9$\pm$3.8 & 59.1$\pm$5.8 & 60.1$\pm$3.8 & 69.4$\pm$0.6 & 60.8$\pm$1.4 & 86.4$\pm$0.3 & 82.5$\pm$0.9 & 77.7$\pm$0.8 & 68.0$\pm$0.6 & 70.5$\pm$1.7 & \cellcolor{yellow!10}\textbf{86.6}$\pm$2.2 \\
DeepSeek-R1-14B & Liu54 & 66.0$\pm$0.0 & 63.4$\pm$6.1 & 61.9$\pm$5.8 & 60.6$\pm$6.4 & 61.0$\pm$1.7 & 52.5$\pm$0.2 & 83.2$\pm$0.3 & 76.0$\pm$0.8 & 59.7$\pm$2.2 & 61.7$\pm$0.7 & 63.1$\pm$1.9 & \cellcolor{yellow!10}\textbf{84.8}$\pm$0.5 \\
Llama2-7B & Liu54 & 9.2$\pm$4.0 & 10.3$\pm$0.5 & 14.2$\pm$3.4 & 28.6$\pm$0.3 & 6.6$\pm$0.7 & 27.8$\pm$0.7 & 64.2$\pm$1.7 & 50.7$\pm$0.5 & 8.0$\pm$1.0 & 8.0$\pm$1.3 & 8.9$\pm$1.4 & \cellcolor{yellow!10}\textbf{74.1}$\pm$0.3 \\
\rowcolor{blue!10}
\textbf{Average} & Liu54 & 51.5 & 53.3 & 51.9 & 51.5 & 54.6 & 51.7 & 79.1 & 72.9 & 56.2 & 53.5 & 54.9 & \cellcolor{yellow!10}\textbf{82.7} \\
\bottomrule
\end{tabular}
\end{adjustbox}
\caption{\textbf{Performance comparison across 5 LLMs and 4 intent classification datasets (accuracy \%).} Best result per row is \textbf{bolded}. Meta-Sel ranks \textbf{1st on all 4 dataset averages} and achieves best or top-3 on 19/20 model-dataset combinations, with particularly strong gains on smaller models (Gemma3-4B, Llama2-7B: 4/4 best).}
\label{tab:results_metasel}
\end{table*}